\documentclass{article}

\PassOptionsToPackage{numbers, compress}{natbib}

\usepackage[preprint]{neurips_2024}

\usepackage[utf8]{inputenc} 
\usepackage[T1]{fontenc}    
\usepackage{hyperref}       
\usepackage{url}            
\usepackage{booktabs}       
\usepackage{amsfonts}       
\usepackage{nicefrac}       
\usepackage{microtype}      
\usepackage{xcolor}         
\usepackage{amsmath}

\usepackage{algorithm}
\usepackage{algcompatible}

\usepackage{cleveref}

\usepackage{graphicx}
\usepackage{tabularx}
\usepackage{subcaption}

\title{Unified Directly Denoising  for Both Variance Preserving and Variance Exploding Diffusion Models}

\author{Jingjing Wang\thanks{Equal contribution}\qquad Dan Zhang$^*$ \qquad Feng Luo \\
\texttt{Clemson University} \\
\texttt{\{jingjiw,dzhang4,luofeng\}@clemson.edu}}

\begin{document}

\maketitle

\begin{abstract}
Previous work has demonstrated that, in the Variance Preserving (VP) scenario, the nascent Directly Denoising Diffusion Models (DDDM) can generate high-quality images in one step while achieving even better performance in multistep sampling. However, the pseudo-LPIPS loss used in DDDM leads to concerns about the bias in assessment. Here, we propose a unified DDDM (uDDDM) framework that generates images in one-step/multiple steps for both Variance Preserving (VP) and Variance Exploding (VE) cases. We provide theoretical proofs of the existence and uniqueness of the model’s solution paths, as well as the non-intersecting property of the sampling paths. Additionally, we propose an adaptive Pseudo-Huber loss function to balance the convergence to the true solution and the stability of convergence process.
Through a comprehensive evaluation, we demonstrate that uDDDMs achieve FID scores comparable to the best-performing methods available for CIFAR-10 in both VP and VE. Specifically, uDDDM achieves one-step generation on CIFAR10 with FID of 2.63 and 2.53 for VE and VP respectively. By extending the sampling to 1000 steps, we further reduce FID score to 1.71 and 1.65 for VE and VP respectively, setting state-of-the-art performance in both cases.
\end{abstract}

\section{Introduction}
Diffusion models have recently emerged as a cutting-edge method in generative modeling for producing high-quality samples in many applications including image, audio \cite{kong2021diffwave,popov2022diffusionbased}, and video generation \cite{latentDiff,Imagen,classifierfree}. These models can be derived from stochastic differential equations (SDEs), that employ a forward process to gradually transform data into Gaussian noise, and a reverse process to reconstruct the initial data from the noise. The two main types of forward SDEs are Variance Preserving (VP) SDE and Variance Exploding (VE) SDE \citep{score, edm}. The former maintains a stationary distribution of isotropic Gaussian and the later mimics Brownian motion but with a deterministic time change that results in an exploding variance.  Despite their effectiveness, the classical diffusion process requires multiple iterative steps to produce a high quality sample.

To tackle the challenge of inefficient inference process, one-step generative models such as the Consistency Models (CMs) \cite{consistency,iCT}, Consistency Trajectory Models (CTMs) \cite{ctm} and TRACT \cite{tract} have been introduced. While the aforementioned innovations in those models considerably enhance sampling efficiency, they still have several drawbacks. For example, one major concern for CMs is the complexity of its hyperparameters. These hyperparameters must be carefully tuned to generate high-quality images, a process that is challenging to optimize and may not generalize well across different tasks or datasets.  
CTM employs adversarial training, which may increase computational cost and reduced training stability, as GANs are notoriously challenging to train \cite{StyleganT}. TRACT is a pure distillation-based model, introducing additional computational overhead during training as it relies on a separate pre-trained diffusion model (the teacher model). Furthermore, the quality of the student model is limited to that of the teacher model.

Preivously, Zhang et. al introduced the  Directly Denoising Diffusion Models (DDDMs,\cite{ours}) , which can generate realistic images in one step while achieving even better performance in multistep sampling in the VP scenario. The DDDM models employ Pseudo-LPIPS loss, which is based on the learned LPIPS \cite{lpips} metrics. However, the Pseudo-LPIPS loss may introduce bias in the evaluation. In this paper, we propose uDDDM, a unified directly denoising diffusion framework for both VP and VE cases. We provide the theoretical proofs for the properties of uDDDMs. Additionally, to overcome the drawbacks of previous Pseudo-LPIPS loss, we have also proposed an adaptive Pseudo-Huber loss function, which can enhance both robustness and sensitivity of uDDDM. We validate uDDDM using the CIFAR-10 benchmark dataset, achieving performance on par with state-of-the-art diffusion models in one-step sampling. Moreover, uDDDM surpasses these models in multi-step sampling scenarios, demonstrating its superior efficacy and robustness.

Our contributions can be summarized as follows:
\begin{itemize}
    \item We introduce uDDDM, which provides a unified framework to support both VP and VE cases. This integrated structure is designed to leverage the strengths of DDDMs, enabling them to perform exceptionally well across different noise schedulers.
    \item We propose an adaptive loss function for training. This function seamlessly integrates two components: a guiding loss and an iterative loss. The guiding loss focuses on minimizing the difference between the predicted and actual data points, while the iterative loss ensures that the model's predictions remain consistent over multiple iterations. By dynamically balancing these components, the adaptive loss function adjusts to the specific needs of the model at different epochs of training. 
    \item We provide theoretical proofs of the existence and uniqueness of the model's solution paths. Furthermore, we establish a non-intersecting sampling path theorem, which ensures that given any two distinct starting points, the multi-step sampling paths will not intersect. This property is crucial for maintaining the diversity and independence of sample paths in complex generative models, enhancing the model's ability to explore the sample space without redundancy.
\end{itemize}

\section{Preliminaries}
\subsection{Variance Exploding (VE) Stochastic Differential Equation (SDE) }

The concept of Variance Exploding (VE) Stochastic Differential Equation (SDE) \cite{score} originates from the Score Matching with Langevin Dynamics (SMLD) \cite{smld}, a method characterized by the use of multiple noise scales ranging from $\sigma_0$ to $\sigma_{N-1}$. In SMLD, a forward Markov chain is utilized where each transition is defined as:

\[\mathbf{x}_i=\mathbf{x}_{i-1}+\sqrt{\sigma_i^2-\sigma_{i-1}^2} \mathbf{z}_{i-1}, 1 \leq i \leq N\]

where \(\mathbf{z}_{i-1} \sim \mathcal{N}(\mathbf{0}, \mathbf{I}), \mathbf{x}_0 \sim p_{\text {data }}\). 

As the number of steps \(N\) goes to infinity, this discrete chain transitions into a continuous stochastic process. In this limit, the discrete states \(\mathbf{x}_i\), the discrete noise scales \(\sigma_i\) and the random variables \(\mathbf{z}_i\) are replaced by continuous functions \(\mathbf{x}_t\), \(\sigma_t\) and \(\mathbf{z}_t\), respectively. The transformation of the updating equation in the continuous limit, with \(\Delta t=\frac{1}{N} \rightarrow 0\), becomes:

\[\mathbf{x}_{t+\Delta t}=\mathbf{x}_t+\sqrt{\sigma_{t+\Delta t}^2-\sigma_t^2} \mathbf{z}_t \approx \mathbf{x}_t+\sqrt{\frac{\mathrm{d}\sigma_t^2}{\mathrm{d} t} \Delta t} \mathbf{z}_t .\]

The above approximation can be expressed as an equality with the following differential form

\[\mathrm{d} \mathbf{x}_t=\sqrt{\frac{\mathrm{d}\sigma_t^2}{\mathrm{d} t}} \mathrm{~d} \mathbf{w}, \quad 0 \leq t \leq T.\]

This is the VE SDE, where \(\mathbf{w}\) denotes the wiener process. 
The diffusion term \(\sqrt{\frac{d \sigma_t^2}{d t}}\) is a monotonically increasing function, typically chosen to be a geometric sequence \cite{smld,score}
by setting $\sigma_t=$ $\sigma_{\min }\left(\frac{\sigma_{\max }}{\sigma_{\min }}\right)^{\frac{t}{T}}$, with $\sigma_{\min } \ll \sigma_{\max }$. 


Thus, the forward process for Variance Exploding case is:

\[
p_{0 t}(\mathbf{x}_t \mid \mathbf{x}_0) =\mathcal{N}\left(\mathbf{x}_t ; \mathbf{x}_0, \sigma_t^2 \mathbf{I}\right), \quad t \in [0,T]
\]


Karras et al.\cite{edm} revisited the noise scheduler of VE SDE and given the following expression: 

\[\sigma_t=\left(\sigma_{\min }^{1 / \rho}+\frac{t-1}{T-1}\left(\sigma_{\max }^{1 / \rho}-\sigma_{\min }^{1 / \rho}\right)\right)^\rho \quad t \in [1,T]\]

Moreover, Karras et al.\cite{edm} stated that as $\rho \rightarrow \infty$, the above equation simplifies to the same geometric sequence used by the original VE SDE. Therefore, this discretization can be viewed as a parametric generalization of the method proposed by Song et al.\cite{score}. 


The reserve SDE \cite{ANDERSON1982,score} of VE can be expressed as: 
\[\mathrm{d} \mathbf{x}_t=\left[-\frac{\mathrm{d}\sigma_t^2}{\mathrm{d} t} \nabla_{\mathbf{x}_t} \log p_t(\mathbf{x}_t)\right] \mathrm{d} t+\sqrt{\frac{\mathrm{d} \sigma_t^2}{\mathrm{~d} t}} \mathrm{~d} \mathbf{w},\]

When we remove the diffusion term, the above reverse SDE can be simplified to the probability flow (PF) ODE as:

\[\mathrm{d} \mathbf{x}_t=\left[-\frac{\mathrm{d}\sigma_t^2}{\mathrm{d} t} \nabla_{\mathbf{x}_t} \log p_t(\mathbf{x}_t)\right] \mathrm{d} t\]

Usually, one can use the exsiting ODE solvers to solve the PF ODE.

\subsection{The framework of DDDM}




Since the numerical solvers of ODE can not avoid the discretization error \cite{bridge}, which restricts the quality of samples when only a small number NFEs are used, 
DDDMs were proposed as an ODE solver-free approach that aims to directly recover the initial point \(\mathbf{x}_0\) of the PF ODE trajectory. This is achieved through an iterative process that refines the estimation.

First, we define \(\mathbf{f}\left(\mathbf{x}_0, \mathbf{x}_t, t\right)\) as the solution of the PF ODE with VP from initial time \(t\) to end time \(0\) : 
\[
\mathbf{f}\left(\mathbf{x}_0, \mathbf{x}_t, t\right):=\mathbf{x}_t+\int_t^0-\frac{1}{2} \beta(s)\left[\mathbf{x}_s-\nabla_{x_s} \log q_s\left(\mathbf{x}_s\right)\right] \mathrm{d} s\]
where $\mathbf{x}_t$ is drawn from \(\mathcal{N}\left(\sqrt{\alpha_t} \mathbf{x}_0,\left(1-\bar{\alpha}_t\right) \mathbf{I}\right) .\)

Next, we defined the function \(\mathbf{F}\left(\mathbf{x}_0, \mathbf{x}_t, t\right)\) as:
\[\mathbf{F}\left(\mathbf{x}_0, \mathbf{x}_t, t\right):=\int_t^0 \frac{1}{2} \beta(s)\left[\mathbf{x}_s-\nabla_{x_s} \log q_s\left(\mathbf{x}_s\right)\right] \mathrm{d} s .\]

Therefore, we have
\[\mathbf{f}\left(\mathbf{x}_0, \mathbf{x}_t, t\right)=\mathbf{x}_t-\mathbf{F}\left(\mathbf{x}_0, \mathbf{x}_t, t\right) .\]
By approximating $\mathbf{f}$, the original image $\mathbf{x}_0$ can be recovered. Let $\mathbf{f}_{\boldsymbol{\theta}}$ be a neural network parameterized function, which is employed to estimate the solution of the PF ODE and thereby recover the original image state at time 0. The predictive model is represented as:
\[
\mathbf{f}_{\boldsymbol{\theta}}\left(\mathbf{x}_0, \mathbf{x}_t, t\right)=\mathbf{x}_t-\mathbf{F}_{\boldsymbol{\theta}}\left(\mathbf{x}_0, \mathbf{x}_t, t\right)
\]
where \(\mathbf{F}_{\boldsymbol{\theta}}\) is the neural network function parameterized with \(\boldsymbol{\theta}\). To achieve a good recovery of the initial state \(\mathbf{x}_0\), it is necessary to ensure \(\mathbf{f}_{\boldsymbol{\theta}}\left(\mathbf{x}_0, \mathbf{x}_t, t\right) \approx \mathbf{f}\left(\mathbf{x}_0, \mathbf{x}_t, t\right)\).

The practical application of DDDM involves an iterative approach, where an initial estimate $\mathbf{x}_0^{(n)}$ is refined iteratively using the update equation:
\[\mathbf{x}_0^{(n+1)}=\mathbf{x}_t-\mathbf{F}_{\boldsymbol{\theta}}\left(\mathbf{x}_0^{(n)}, \mathbf{x}_t, t\right) .\]



\section{A Unified DDDM (uDDDM) Framework}
In this section, we present a unified DDDM framework, which integrate both VP and VE approaches into a single framework. Additionally, we introduce an adaptive loss function designed to enhance the robustness and stability of the training process. 

 


\subsection{The Unified DDDMs}

The PF ODE can be expressed as:
\begin{equation}
\label{eq:2}
\frac{\mathrm{d} \mathbf{x}_t}{\mathrm{~d} t}=-\mathbf{h}\left(\mathbf{x}_0, \mathbf{x}_t, t\right)\end{equation}

where we denote:

\[
\mathbf{h}\left(\mathbf{x}_0, \mathbf{x}_t, t\right) := 
\begin{cases}
\begin{aligned}
\frac{\mathrm{d}\sigma_t^2}{\mathrm{d} t} \nabla_{\mathbf{x}_t} \log p_t(\mathbf{x}_t) & \quad \text{(VE case)} \\
\frac{1}{2} \beta(t)\left[\mathbf{x}_t - \nabla_{x_t} \log q_t\left(\mathbf{x}_t\right)\right] & \quad \text{(VP case)}
\end{aligned}
\end{cases}
\]


 We integrate both sides from initial time \(t\) to final time \(0\) of \Cref{eq:2} and obtain:
 \[\mathbf{f}\left(\mathbf{x}_0, \mathbf{x}_t, t\right)=\mathbf{x}_t-\int_t^0 \mathbf{h}\left(\mathbf{x}_0, \mathbf{x}_s, s\right) \mathrm{d} s\]

For stable training purpose in VE case, 
we rewrite the \(\mathbf{f}\left(\mathbf{x}_0, \mathbf{x}_t, t\right)\) as a combination of \(\mathbf{x}_t\) and a function \(\mathbf{F}\).This approach helps to balance the contributions of the current state and the function, thereby improving the robustness of the training process.

\begin{equation}
\label{eq:3}
\begin{aligned}
\mathbf{f}\left(\mathbf{x}_0, \mathbf{x}_t, t\right) & =\mathbf{x}_t-\underbrace{\int_t^0 \mathbf{h}\left(\mathbf{x}_0, \mathbf{x}_s, s\right) \mathrm{d} s}_{\mathbf{F}\left(\mathbf{x}_0, \mathbf{x}_t, t\right)} & (\text{VP case}) \\
& =\kappa(\sigma_t) \mathbf{x}_t+(1-\kappa(\sigma_t)) \mathbf{x}_t-\int_t^0 \mathbf{h}\left(\mathbf{x}_0, \mathbf{x}_s, s\right) \mathrm{d} s
\\& =\kappa(\sigma_t) \mathbf{x}_t+(1-\kappa(\sigma_t)) \underbrace{\left[\mathbf{x}_t-\frac{1}{1-\kappa(\sigma_t)} \int_t^0 \mathbf{h}\left(\mathbf{x}_0, \mathbf{x}_s, s\right) \mathrm{d} s\right]}_{\mathbf{F}\left(\mathbf{x}_0, \mathbf{x}_t, t\right)} & (\text{VE case}) \\
\end{aligned}
\end{equation}

From \Cref{eq:3}, we can obtain a unified expression for \(\mathbf{f}(\mathbf{x}_0, \mathbf{x}_t, t)\):
\begin{equation*}
    \mathbf{f}(\mathbf{x}_0, \mathbf{x}_t, t) = a(\sigma_t) \mathbf{x}_t + b(\sigma_t) \mathbf{F}(\mathbf{x}_0, \mathbf{x}_t, t),
\end{equation*}

where for VP case, \( a(\sigma_t)=1,  b(\sigma_t)=-1, \mathbf{x}_t \sim \mathcal{N}\left(\sqrt{\bar{\alpha}_t} \mathbf{x}_0,\left(1-\bar{\alpha}_t\right) \mathbf{I}\right)\) and \(\bar{\alpha}_t=\prod_{s=1}^t\left(1-\beta_s\right)\). For VE model, \(a(\sigma_t)=\kappa(\sigma_t)\),  \(b(\sigma_t)=1-\kappa(\sigma_t)\), \(\mathbf{x}_t \sim \mathcal{N}(\mathbf{x}_0,\sigma_t^2 \mathbf{I})\). There are a few choices for the design of \(\kappa(\sigma_t)\), such as \(\kappa(\sigma_t) = \frac{\sigma_{data}}{\sigma_t+\sigma_{data}}\), \(\kappa(\sigma_t) = \frac{\sigma_{data}^2}{\sigma_t^2+\sigma_{data}^2}\). We set \(\kappa(\sigma_t) = \frac{\sigma_{\text{min}}}{\sigma_t}\) in this work.

To estimate \(\mathbf{x}_0\), we predict:

\[\mathbf{f}_{\boldsymbol{\theta}}\left(\mathbf{x}_0, \mathbf{x}_t, t\right) = a\left(\sigma_t\right) \mathbf{x}_t+b\left(\sigma_t\right) \mathbf{F}_{\boldsymbol{\theta}}\left(\mathbf{x}_0, \mathbf{x}_t, t\right)\]

and employ an iterative process for this prediction:
\[
\mathbf{x}_0^{(n+1)}=\mathbf{f}_{\boldsymbol{\theta}}\left(\mathbf{x}_0^{(n)}, \mathbf{x}_t, t\right)
\]
Through this iterative procedure, we are able to progressively approach the actual starting point by improving the estimate of \(\mathbf{x}_0\) at each training epoch \(n\). The refinement equation becomes:
\[
\mathbf{x}_0^{(n+1)}=a\left(\sigma_t\right) \mathbf{x}_t+b\left(\sigma_t\right) \mathbf{F}_{\boldsymbol{\theta}}\left(\mathbf{x}_0^{(n)}, \mathbf{x}_t, t\right)
\]
To effectively minimize the discrepancy between the iteratively estimated \(\mathbf{x}_0^{(n)}\) and the true initial state \(\mathbf{x}_0\), we employ a specific loss function during the training of our model. This loss function ensures that the neural network learns to produce accurate estimates of the initial state by reducing the error between the predicted and actual values.

\subsection{The Adaptive Loss Function}

To balance the convergence to the true solution and the stability of convergence process, we propose an adaptive loss function:
\begin{equation}
    \label{eq:loss}
    \mathcal{L}_{\text{uDDDM}}^{(n)}(\boldsymbol{\theta}):= \frac{1}{n+1}\mathcal{L}_{\text{Guide}}^{(n)}(\boldsymbol{\theta}) + (1-\frac{1}{n+1})\mathcal{L}_{\text{Iter}}^{(n)}(\boldsymbol{\theta})
\end{equation}
where
\begin{gather*}
    \mathcal{L}_{\text{Guide}}^{(n)}(\boldsymbol{\theta}):=\mathbb{E}_{\mathbf{x}_0,\mathbf{x}_t,t}\left[d\left(\mathbf{f}_{\boldsymbol{\theta}}\left(\mathbf{x}_0^{(n)}, \mathbf{x}_t, t\right), \mathbf{x}_0\right)\right]\\
    \mathcal{L}_{\text{Iter}}^{(n)}(\boldsymbol{\theta}):=\mathbb{E}_{\mathbf{x}_0,\mathbf{x}_t,t} \left[d\left(\mathbf{f}_{\boldsymbol{\theta}}\left(\mathbf{x}_0^{(n)}, \mathbf{x}_t, t\right), \mathbf{x}_0^{(n)}\right)\right]
\end{gather*}




  In both loss functions, \(t\) is sampled from a uniform distribution over the integer set \(\left[1,2, \cdots, T\right]\), \(\mathbf{x}_0 \sim p_{\text {data }}\), \(\mathbf{x}_t \sim \mathcal{N}\left(\sqrt{\bar{\alpha}_t} \mathbf{x}_0,\left(1-\bar{\alpha}_t\right) \mathbf{I}\right) \) or \(\mathcal{N}\left(\mathbf{x}_0, \sigma_t^2 \boldsymbol{I}\right)\) and \(n\) denotes training epoch starting from \(0\). \(d(\cdot, \cdot)\) is a metric function satisfies that for all vectors \(\mathbf{x}\) and \(\mathbf{y}\), \(d(\mathbf{x}, \mathbf{y}) \geq 0\) and \(d(\mathbf{x}, \mathbf{y})=0\) if and only if \(\mathbf{x}=\mathbf{y}\). Therefore, commonly used metrics such as \(\mathcal{L}_1\) or \(\mathcal{L}_2\) can be utilized. We will discuss our choice of \(d(\cdot, \cdot)\) later.

This loss function is designed to measure the performance of our proposed method in two distinct aspects. 

\textbf{Convergence to the True Solution:} The first term \(\mathcal{L}_{\text{Guide}}^{(n)}(\theta)\) is the guiding loss, which decreases the weight of the deviation of  \(\mathbf{f}_\theta(\mathbf{x}_0^{(n)}, \mathbf{x}_t, t)\) from the true value \(\mathbf{x}_0\) as \(n\) increases. This implies that in the early stage of training process, model estimation \(\mathbf{x}_0^{(n)}\) may be far away from ground truth. Therefore, exact alignment with the true value will be emphasized. 

 \textbf{Stability of the Iteration:} The second term \(\mathcal{L}_{\text{Iter}}^{(n)}(\theta)\) denotes the iterative loss, which measures the self-consistency of the iteration as it progresses. It assesses how close \(\mathbf{f}_\theta\left(\mathbf{x}_0^{(n)}, \mathbf{x}_t, t\right)\) is to \(\mathbf{x}_0^{(n)}\), with increasing importance given to this term as \(k\) increases. This reflects a growing emphasis on the iteration's stability and self-consistency as it proceeds, which is critical for the convergence of our proposed methods.

The loss formulation effectively balances between guiding the iteration towards the true fixed point and ensuring the method stabilizes. Early in the training process, the emphasis is more on aligning with the true solution rather than stabilizing the method, which can be particularly advantageous when \(\mathbf{x}_0\) is not well-approximated initially.

As \(n\) increases, reducing the weight on the first term allows the iterations to focus more on refining the solution to ensure it is a fixed point, rather than merely approximating the true solution. This shift is crucial for the practical implementation of iterative methods, as it acknowledges the dual requirements of convergence and stability, which are often at odds in numerical computations.

This approach is particularly well-suited for problems where the true solution may be difficult to approach directly due to complexities in the function $F$ or the initial conditions. By adjusting the focus from accuracy towards stability as iterations progress, the method can achieve a more reliable convergence, making it robust in various scenarios.




\textbf{Metric function.} Inspired by \cite{iCT}, we adopt the Pseudo-Huber metric family \cite{PH} for function $d(\cdot,\cdot)$ in \Cref{eq:loss}, defined as
\begin{equation}
    d(\boldsymbol{x},\boldsymbol{y}) = \sqrt{\|\boldsymbol{x}-\boldsymbol{y}\|_2^2+c^2}-c
\end{equation}
where $c$ is an adjustable hyperparamter. The Pseudo-Huber metric is more robust to outliers compared to the squared \(\mathcal{L}_2\) metric because it imposes a smaller penalty for large errors, while still behaving similarly to the squared \(\mathcal{L}_2\) metric for smaller errors. Additionally, the Pseudo-Huber metric is unbiased in evaluation compared to LPIPS \cite{lpips}. This is because both LPIPS and the Inception network used for FID employ ImageNet \cite{imagenet}. The potential leakage of ImageNet features from LPIPS could result in inflated FID scores. We set $c=0.00014$ for VP and $c=0.00015$ for VE respectively.

\subsection {Theoretical Justifications of uDDDM}
The theoretical justifications presented for uDDDM demonstrate the robustness of the model under certain mathematical conditions, ensuring its reliability in practical applications.  

Suppose \(\mathbf{F}_{\boldsymbol{\theta}}\left(\mathbf{x}_0^{(n)}, \mathbf{x}_t, t\right)\) is twice continuously differentiable with bounded first and second derivatives and denote \(\mathbf{h}_{\boldsymbol{\theta}}\left(\mathbf{x}_0^{(n)}, \mathbf{x}_t, t\right) :=\frac{\mathrm{d} \mathbf{F}_{\boldsymbol{\theta}}\left(\mathbf{x}_0^{(n)}, \mathbf{x}_t, t\right)}{\mathrm{d} t}\) or \(\frac{\left(1-\kappa\left(\sigma_t\right)\right)\mathrm{d} \mathbf{F}_{\boldsymbol{\theta}}\left(\mathbf{x}_0^{(n)}, \mathbf{x}_t, t\right)}{\mathrm{d} t}\) for VP and VE, respectively.

The initial value problem (IVP) of the PF ODE can be expressed as:

\begin{equation}
\label{ivp:1}
\left\{
\begin{aligned}
    \frac{\mathrm{d} \mathbf{x}_s}{\mathrm{~d} s}&=-\mathbf{h}_{\boldsymbol{\theta}}\left(\mathbf{x}_0^{(n)}, \mathbf{x}_s, s\right) && \quad s \in [0, t] \\
    \mathbf{x}_t &= \hat{\mathbf{x}}_t
\end{aligned}
\right.
\end{equation}

if putting \(\Tilde{\mathbf{x}}_s:=\mathbf{x}_{t-s}\), we get 
\begin{equation}
\label{ivp:2}
\left\{
\begin{aligned}
    \frac{\mathrm{~d} \Tilde{\mathbf{x}}_s}{\mathrm{~d} s}&=\mathbf{h}_{\boldsymbol{\theta}}\left(\Tilde{\mathbf{x}}_t^{(n)}, \Tilde{\mathbf{x}}_s, s\right) && \quad s \in [0, t] \\
    \Tilde{\mathbf{x}}_0 &= \hat{\mathbf{x}}_t
\end{aligned}
\right.
\end{equation}

The IVP \ref{ivp:1} and \ref{ivp:2} are equivalent and can be used interchangeably. 

\textbf{Theorem 1 (Uniqueness)}: Suppose we have the initial value problem \ref{ivp:2},
where, \(\hat{\mathbf{x}}_t\) is a given initial condition at time \(0\) for \(\Tilde{\mathbf{x}}_0\). 
Assume that there exists \(L>0\), such that the following Lipschitz condition holds: 
\begin{equation}
\label{eq:lip}
\|\mathbf{h}_{\boldsymbol{\theta}}\left(\Tilde{\mathbf{x}}_t^{(n)}, \Tilde{\mathbf{x}}_s, s\right)-\mathbf{h}_{\boldsymbol{\theta}}\left(\Tilde{\mathbf{y}}_t^{(n)}, \Tilde{\mathbf{y}}_s, s\right)\|_2\leq L\|\Tilde{\mathbf{x}}_s-\Tilde{\mathbf{y}}_s\|_2 
\end{equation}
for all \(s \in[0, t]\) and \(\Tilde{\mathbf{x}}_s, \Tilde{\mathbf{y}}_s \in \mathbb{R}^D\). Then there exists at most one function \(\mathbf{x}_t\) which satisfies the initial value problem.

Theorem 1 assures that given an initial state, the evolution of the model is deterministic and predictable within the bounds of the Lipschitz condition. 







\textbf{Theorem 2}:
 If the loss function \(\mathcal{L}_{\text{uDDDM}}^{(n)}(\boldsymbol{\theta}) \rightarrow 0\) as \(n \rightarrow \infty\), it can be shown that as \(n \rightarrow \infty\),

\[\mathbf{f}_{\boldsymbol{\theta}}\left(\mathbf{x}_0^{(n)}, \mathbf{x}_t, t\right) \rightarrow \mathbf{x}_0 .\]

Theorem 2 extends aforementioned assurance, linking the convergence of the loss function to give existence of the solution. That is, the iterative solution will recover the true \(\mathbf{x}_0\) when the neural network is sufficiently trained. 

\textbf{Theorem 3 (Non-Intersection)}: Suppose the neural network is sufficiently trained, \(\boldsymbol{\theta}^{*}\) obtained such that \(\mathbf{f}_{\boldsymbol{\theta}^*}(\mathbf{x}_0^{(n)},\mathbf{x}_t, t) \equiv \mathbf{f}(\mathbf{x}_0,\mathbf{x}_t, t)\) for any \(t \in [0, T]\) and \(\mathbf{x}_0\) sampled from \(p_{\text{data}}\), and \(\mathbf{h}_{\boldsymbol{\theta}}\left(\tilde{\mathbf{x}}_t^{(n)}, \tilde{\mathbf{x}}_s, s\right)\) meets Lipschitz condition (\Cref{eq:lip}) 

Then for any \(t \in[0, T]\), the mapping \(\mathbf{f}_{\boldsymbol{\theta}^*}(\mathbf{x}_0^{(n)},\mathbf{x}_t, t): \mathbb{R}^D \rightarrow \mathbb{R}^D\) is bi-Lipschitz. Namely, for any \(\mathbf{x}_t, \mathbf{y}_t \in \mathbb{R}^D\)

\[
e^{-Lt}\left\|\mathbf{x}_t-\mathbf{y}_t\right\|_2 \leq \left\|\mathbf{f}_{\boldsymbol{\theta}^*}(\mathbf{x}_0^{(n)},\mathbf{x}_t, t)-\mathbf{f}_{\boldsymbol{\theta}^*}(\mathbf{y}_0^{(n)},\mathbf{y}_t, t)\right\|_2 \leq e^{Lt}\left\|\mathbf{x}_t-\mathbf{y}_t\right\|_2 .
\]

This implies that if given two different starting point, say \(\mathbf{x}_T\neq \mathbf{y}_T\), by the bi-Lipschitz above, it can be conculde that \(\mathbf{f}_{\boldsymbol{\theta}^*}(\mathbf{x}_0^{(n)},\mathbf{x}_T, T)\neq \mathbf{f}_{\boldsymbol{\theta}^*}(\mathbf{y}_0^{(n)},\mathbf{y}_T, T)\) i.e., \(\mathbf{x}_0^{(n+1)} \neq \mathbf{y}_0^{(n+1)}\), which indicate the reverse path of uDDDM does not intersect.


The proof of Theorems presented in \Cref{App}.

\subsection{Training and Sampling with uDDDM}
\label{sec:training}
\textbf{Training}. Each data sample \(\mathbf{x}_0\) is chosen randomly from the dataset, following the probability distribution \(p_{\text {data }}(\mathbf{x}_0)\). This initial data point forms the basis for generating a trajectory. Next, we randomly sample a \(t \sim \mathcal{U}[1, T]\), and obtain 
its noisy variant \(\mathbf{x}_t\) accordingly.
we play the reparameterization trick to rewrite $\mathbf{x}_t = \mathbf{x}_0+\sigma_t\boldsymbol{\epsilon}$ for VE and \(\mathbf{x}_t = \sqrt{\bar{\alpha}_t} \mathbf{x}_0 + \sqrt{1-\bar{\alpha}_t}\boldsymbol{\epsilon}\) for VP, where $\boldsymbol{\epsilon} \sim \mathcal{N}(\mathbf{0}, \mathbf{I})$. For current training epoch \(n\), our model takes noisy data \(\mathbf{x}_t\) and timestep \(t\), as well as the corresponding estimated target from previous epoch \(x_0^{(n-1)}\) as inputs, predicts a new approximation \(x_0^{(n)}\), which will be utilized in the next training epoch. uDDDM is trained by minimizing the loss following Eq. \ref{eq:loss}.
The full procedure of training uDDDM is summarized in Algorithm \ref{algo:1}.

\textbf{Sampling}. The generation of samples is facilitated through the use of a well-trained uDDDM, denoted as \(\mathbf{f}_{\boldsymbol{\theta}}(\cdot, \cdot)\). The process begins by drawing from the initial Gaussian distribution, where both \(\mathbf{x}_0^{(0)}\) and \(\mathbf{x}_T\) are sampled from \(\mathcal{N}\left(\mathbf{0}, \boldsymbol{I}\right)\). $\mathbf{x}_T$ will be scaled by $\sigma_\text{max}$ for VE case. Subsequently, these noise vectors and embedding of $T$ are passed through the uDDDM model to obtain 
\(\mathbf{x}_0^{\text{est}}=
\mathbf{f}_{\boldsymbol{\theta}}\left(\mathbf{x}_0^{(0)}, \mathbf{x}_T, T\right)
\). This approach is noteworthy for its efficiency, as it requires only a single forward pass through the model. Our model also supports a multistep sampling procedure for enhanced sample quality. Detail can be found in Algorithm \ref{algo:2}.

\begin{algorithm}[t]
   \caption{Training}
\begin{algorithmic}
   \STATE {\bfseries Input:} image dataset $D,$ $T$, model parameter $\boldsymbol\theta$, initialize $\mathbf{x}_0^{(0)} \sim \mathcal{N}(\mathbf{0}, \mathbf{I})$,
    epoch $n \leftarrow 0$
   \REPEAT
   \STATE Sample $\mathbf{x}_0 \sim D$,$t \sim \mathcal{U}\left[1, T\right]$ and $\boldsymbol{\epsilon} \sim \mathcal{N}(\mathbf{0}, \mathbf{I})$
   \IF{VE}
   \STATE $\mathbf{x}_t = \mathbf{x}_0+\sigma_t\boldsymbol{\epsilon}$
   \ELSE
   \STATE $\mathbf{x}_t = \sqrt{\bar{\alpha}_t} \mathbf{x}_0+\sqrt{1-{\bar\alpha}_t} \boldsymbol{\epsilon} $
   \ENDIF
   \STATE$\mathbf{x}_0^{(n+1)} \leftarrow \mathbf{f}_{\boldsymbol\theta}\left(\mathbf{x}_{0}^{(n)}, \mathbf{x}_t, t\right)$ 
   \STATE $\mathcal{L}_{\text{uDDDM}}^{(n)}(\boldsymbol{\theta})\leftarrow \frac{1}{n+1}[d(\mathbf{f}_\theta(\mathbf{x}_0^{(n)}, \mathbf{x}_t, t), \mathbf{x}_0)] + (1-\frac{1}{n+1})[d(\mathbf{f}_\theta(\mathbf{x}_0^{(n)}, \mathbf{x}_t, t), \mathbf{x}_0^{(n)})]$
   
   \STATE $\boldsymbol{\theta} \leftarrow \boldsymbol{\theta}-\eta \nabla_{\boldsymbol{\theta}} \mathcal{L}\left(\boldsymbol{\theta}\right)$
   \STATE $n \leftarrow n+1$
   \UNTIL{convergence}
\end{algorithmic}
\label{algo:1}
\end{algorithm}

\begin{algorithm}[t]
   \caption{Sampling}
\begin{algorithmic}
   \STATE {\bfseries Input:} $T$, trained model parameter $\boldsymbol\theta$, sampling step $s$, initialize
   $\mathbf{x}_0^{(0)} \sim \mathcal{N}(\mathbf{0}, \mathbf{I})$, $\mathbf{x}_T \sim \mathcal{N}(\mathbf{0}, \mathbf{I})$
   \IF{VE}
   \STATE $\mathbf{x}_T = \sigma_\text{max}\mathbf{x}_T$
   \ENDIF
   \FOR{$n=0$ {\bfseries to} $s-1$}
   \STATE $\mathbf{x}_0^{(n+1)} \leftarrow \mathbf{f}_{\boldsymbol\theta}\left(\mathbf{x}_{0}^{(n)}, \mathbf{x}_T, T\right)$ 
   \ENDFOR
   \STATE \textbf{Output:} $\mathbf{x}_0^{(n+1)}$
   
\end{algorithmic}
\label{algo:2}
\end{algorithm}

\section{Experiments}
To evaluate our method for image generation, we train several uDDDMs on CIFAR-10 \cite{CIFAR} benchmark their performance with
competing methods in the literature. Results are
compared according to Frechet Inception Distance (FID, \citep{FID}), which is computed between 50K generated samples and the whole training set. We also employ Inception Score (IS,
\cite{IS}) to measure sample quality.

\begin{minipage}[t]{\textwidth}
\begin{minipage}{0.5\textwidth}
\setlength\tabcolsep{0.01cm}
    \centering
    \captionof{table}{Comparing the quality of unconditional samples on CIFAR-10}
    \fontsize{8.5pt}{8.5pt}\selectfont
    \begin{tabular}{lccc}
    \toprule
         Method & NFE($\downarrow$) & FID($\downarrow$) & IS($\uparrow$) \\
         \\
         \multicolumn{4}{l}{\textbf{Fast samplers \& distillation for diffusion models}} \\
         \toprule
         DDIM \cite{ddim}& 10 & 13.36 & \\
         DPM-solver-fast \cite{dpm-solver}& 10 & 4.70 \\
         3-DEIS \cite{3deis}& 10 & 4.17 \\
         UniPC \cite{unipc}& 10 & 3.87 \\
         DFNO (LPIPS) \cite{DFNO} &1 &3.78 \\
         2-Rectified Flow \cite{rectified} &1 &4.85 &9.01\\
         Knowledge Distillation \cite{knowledge} & 1 & 9.36\\
         TRACT \cite{tract}&1 &3.78 \\
         &2 &3.32 \\
         Diff-Instruct \cite{instruct}&1 &4.53 &9.89 \\
         CD (LPIPS) \cite{consistency} &1 &3.55 &9.48 \\
         &2 &2.93 &9.75 \\
         
         \textbf{Direct Generation} \\
         \toprule
         Score SDE \cite{score}&2000 &2.38 &9.83 \\
         Score SDE (deep) \cite{score}  &2000 &2.20 &9.89 \\
DDPM \cite{ddpm} &1000 &3.17 &9.46 \\
LSGM  \cite{LSGM}&147 &2.10 \\
PFGM  \cite{pfgm} &110 &2.35 &9.68 \\
EDM   \cite{edm}&35  &2.04 &9.84 \\
EDM-G++ \cite{edmG}&35 &1.77 \\
NVAE  \cite{nvae}&1 &23.5 &7.18 \\
BigGAN  \cite{bigGAN}&1 &14.7 &9.22 \\
StyleGAN2 \cite{style2}&1 &8.32 &9.21 \\
StyleGAN2-ADA \cite{stylegan2-ada}  &1 &2.92 &9.83 \\
CT (LPIPS) \cite{consistency}&1 &8.70 &8.49 \\
&2 &5.83 &8.85 \\
iCT \cite{iCT} &1 &2.83 &9.54 \\
&2 &2.46 &9.80 \\
iCT-deep \cite{iCT} &1 & 2.51 & 9.76 \\
&2 &2.24 &9.89 \\

\textbf{uDDDM(VE)} &1 &2.91 &9.56 \\
&2 &2.68 &9.75 \\
&1000 &1.89 &9.93 \\

\textbf{uDDDM(VP)} &1 &2.84 &9.56 \\
&2 &2.50 &9.76 \\
&1000 &1.73 &9.94 \\

\textbf{uDDDM(VE-deep)} &1 &2.63 &9.77 \\
&2 &2.35 &9.88 \\
&1000 &1.71 &9.95 \\

\textbf{uDDDM(VP-deep)} &1 &2.53 &9.80 \\
&2 &2.21 &9.90 \\
&1000 &1.65 &9.95 \\

\bottomrule
         
    \end{tabular}
    
    \label{tab:cifar}
\end{minipage}
\hfill
\begin{minipage}{0.45\textwidth}
    
    \includegraphics[scale=0.7]{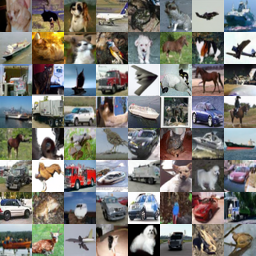}
    \captionof{figure}{One-step samples from uDDDM-VE-deep}
    \label{fig:ve_deep_1}
    \vspace{0.5cm}
    \includegraphics[scale=0.7]{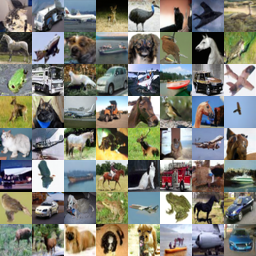}
    \captionof{figure}{One-step samples from uDDDM-VP-deep}
    \label{fig:vp_deep_1}
\end{minipage}
\end{minipage}

\subsection{Implementation Details}
\label{sec:implementation}
\textbf{Architecture} We use the U-Net architecture from ADM \cite{ADM} for both VP and VE setting. Specifically, we use a base channel dimension of 128, multiplied by 1,2,2,2 in 4 stages and 3 residual blocks per stage. Dropout \cite{dropout} of 0.3 is utilized for this task. Following ADM, we employ cross-attention modules not only at the 16x16 resolution but also at the 8x8 resolution, through which we incorporate the conditioning image $\mathbf{x}_0^{(n)}$ into the network.We also explore deeper variants of these architectures by doubling the number of blocks at each resolution, which we name uDDDM-deep. All models on CIFAR-10 are unconditional.

\textbf{Other settings} We use Adam optimizer for all of our experiments.
For VP model, we set $T=8000$ and use a linear variance schedule from $\beta_0 = 1.5\times 10^{-3}$ to $\beta_T = 2.0\times 10 ^{-2}$. For VE model, we use $T=1000$ and set $\sigma_{\text{min}}=0.01$ and  $\sigma_{\text{max}}=50$ respectively. For both settings, we train the models for 400k iterations with a constant learning rate of 0.0002 and batch size of 1024. We use an exponential moving average (EMA) of the weights during training with a decay factor of 0.9999 for all the experiments. All models are trained on 8 Nvidia A100 GPUs.

\subsection{Comparison to SOTA} We compare our model against state-of-the-art generative models on CIFAR-10. Quantitative results are summarized in  \Cref{tab:cifar}. Our findings reveal that uDDDMs exceed previous distillation diffusion models and methods that require advanced sampling procedures in both one-step and two-step generation, which breaks the reliance on the well-pretrained diffusion models and simplifies the generation workflow. Moreover, our model demonstrates performance comparable to numerous leading generative models for both VE and VP settings. Specifically, baseline uDDDM obtains FIDs of 2.91 and 2.84 for one-step generation in VE and VP respetively, both results exceed that of StyleGAN2-ADA \cite{stylegan2-ada}. With 1000-step sampling, our VP baseline further reduces FID to 1.73, outperforming state-of-the-art method \cite{edmG}. For deeper architecture, our model achieves one-step generation with FID of 2.63 and 2.53 for VE and VP respectively. Additionally, VP-deep outperforms the leading model iCT-deep \cite{iCT} on two-step generation. With 1000-step sampling, VE-deep and VP-deep push FID to 1.71 and 1.65 respectively, setting state-of-the-art performance in both cases.

\section{Related Works}
Several one-step generative models have been proposed to improve the effectiveness of inference process, including the Consistency Models (CMs) \cite{consistency,iCT}, Consistency Trajectory Models (CTMs) \cite{ctm} and TRACT \cite{tract}. CMs are available in two variations: Consistency Distillation (CD) and Consistency Training (CT).  CMs focus on enforcing the self-consistency property and share a deep mathematical connection with Score-based generative models (SGMs) \cite{score} and EDMs \cite{edm}. It is accomplished by tracing any point along the path of the same Probability Flow Ordinary Differential Equation (PF ODE) back to the starting point of the trajectory. CTMs generalize CMs and SGMs, and are proposed to address challenges in score-based and distillation sampling methods, by combining decoding strategies for sampling either through SDE/ODE or direct jumps along the PF ODE trajectory. TRACT is a distillation-only method that demonstrates how separating the diffusion trajectory into stages can enhance performance. Its methodology is similar to consistency distillation. 
These models keep the advantages of multi-step sampling for better sample quality while drastically decrease inference time by generating samples in a single step. 

\section{Discussion and Limitations}
\label{sec:limitations}
In this paper, we present uDDDM, a unified framework that can handle both VP and VE scenarios, showcasing its versatility and robustness. The introduction of an adaptive loss function plays a crucial role in this framework, enabling the generation of images with FID scores that are comparable to or even surpass existing benchmarks. Additionally, the theoretical foundations of uDDDM are solidified through a series of proofs, further validating the proposed model's properties and effectiveness. Experimental results on CIFAR10 underscore the potential of uDDDM in advancing the state-of-the-art in the field of diffusion models.

Since uDDDM keeps track of $\mathbf{x}_0^{(n)}$ for each sample in the dataset, there will be additional memory consumption during training. Specifically, it requires extra 614MB for CIFAR10. Although it can be halved by using FP16 data type, such memory requirement might still be a challenge for larger dataset or dataset with high-resolution images. One solution is to store \(\mathbf{x}_0^{(n)}\) in a buffer or on disk instead of on the GPU. However, this approach will introduce additional overhead during training due to the need to transfer data back to the GPU.

We have observed that our VE model consistently underperforms compared to the VP model across all experiments. We hypothesize that this performance gap may be attributed to suboptimal hyperparameters in the loss function. Specifically, our current choice of the hyperparameter \( c \) for the VE model is derived from the corresponding value used for the VP model. However, the optimal value for \( c \) in the VE model might require further fine-tuning to achieve better performance.
Additionally, we recognize that different noise schedulers can significantly impact the model's performance. The noise scheduling strategy plays a crucial role in the training dynamics and final performance of the model. We plan to investigate more complex schedulers in future work.

\bibliography{example_paper}
\bibliographystyle{plain}

\newpage
\appendix

\section{Appendix / supplemental material}\label{App}

Proof of \textbf{Theorem 1}:

Suppose that there are two solutions \(\tilde{\mathbf{x}}_t\) and \(\tilde{\mathbf{y}}_t\) to the initial value problem with the same initial condition \(\tilde{\mathbf{x}}_0=\tilde{\mathbf{y}}_0=\hat{\mathbf{x}}_t\).

Let \(\mathbf{z}_t=\tilde{\mathbf{x}}_t-\tilde{\mathbf{y}}_t\). Then The function \(\mathbf{z}_t\) satisfies the differential equation:
\[\frac{\mathrm{d} \mathbf{z}_s}{\mathrm{~d} s}=\frac{\mathrm{d} \tilde{\mathbf{x}}_s}{\mathrm{~d} s}-\frac{\mathrm{d} \tilde{\mathbf{y}}_s}{\mathrm{~d} s}=\mathbf{h}_{\boldsymbol{\theta}}\left(\tilde{\mathbf{x}}_t^{(n)}, \tilde{\mathbf{x}}_s, s\right)-\mathbf{h}_{\boldsymbol{\theta}}\left(\tilde{\mathbf{y}}_t^{(n)}, \tilde{\mathbf{y}}_s, s\right) \]

Using the given Lipschitz condition:
\[\left\|\mathbf{h}_{\boldsymbol{\theta}}\left(\tilde{\mathbf{x}}_t^{(n)}, \tilde{\mathbf{x}}_s, s\right)-\mathbf{h}_{\boldsymbol{\theta}}\left(\tilde{\mathbf{y}}_t^{(n)}, \tilde{\mathbf{y}}_s, s\right)\right\|_2 \leq L\left\|\tilde{\mathbf{x}}_s-\tilde{\mathbf{y}}_s\right\|_2 \]
we have:
\[
\left\|\frac{d \mathbf{z}_s}{d s}\right\|_2 \leq L\left\|\mathbf{z}_s\right\|_2
\]

Grönwall's inequality states that if $\beta$ and $u$ be real-valued continuous functions defined on \([0,t]\) and $u$ is differentiable in the interior \((0,t)\) and satisfies the differential inequality
\[
u^{\prime}(s) \leq \beta(s) u(s), \quad s \in (0,t)
\]

then \(u\) is bounded by the solution of the corresponding differential equation
\[
u(s) \leq u(a) \exp \left(\int_0^s \beta(v) \mathrm{d} v\right),  \quad s \in (0,t)
\]

Let \(u(s)=\left\|\mathbf{z}_s\right\|_2\). Applying Grönwall's inequality, we obtain
\[
\left\|\mathbf{z}_s\right\|_2 \leq\left\|\mathbf{z}_0\right\|_2 \exp (L t)
\]
Since $\mathbf{z}_0=\Tilde{\mathbf{x}}_0-\Tilde{\mathbf{y}}_0=\hat{\mathbf{x}}_t-\hat{\mathbf{x}}_t=0$, leading to
\[
\left\|\mathbf{z}_0\right\|_2=0
\]

Therefore, for all \(s \in[0, t]\) :
\[
\left\|\mathbf{z}_s\right\|_2 \leq 0 \exp (L s)=0
\]
which implies \(\mathbf{z}_s=0\) for all \(s \in[0, t]\).

Hence,  we have \(\mathbf{x}_t=\mathbf{y}_t \) for all \( t \in[0, T]\), indicating that there is at most one solution to the initial value problem. 

Proof of \textbf{Theorem 2}

As \( n\) sufficiently large,  \(\mathcal{L}_{\mathrm{uDDDM}}^{(n)}(\boldsymbol{\theta}) \rightarrow 0\), we have

\[\left(1-\frac{1}{n+1}\right) \mathcal{L}_{\text {Iter }}^{(n)}(\boldsymbol{\theta}) \rightarrow 0\]implies:
$$
\mathbb{E}_{\mathbf{x}_0, \mathbf{x}_t, t}\left[d\left(\mathbf{f}_{\boldsymbol{\theta}}\left(\mathbf{x}_0^{(n)}, \mathbf{x}_t, t\right), \mathbf{x}_0^{(n)}\right)\right] = 0
$$

which means that \[d\left(\mathbf{f}_{\boldsymbol{\theta}}\left(\mathbf{x}_0^{(n)}, \mathbf{x}_t, t\right), \mathbf{x}_0^{(n)}\right)=0\]
since \(d(\mathbf{x}, \mathbf{y})=0 \Leftrightarrow \mathbf{x}=\mathbf{y}\), it further implies:

\[\mathbf{f}_{\boldsymbol{\theta}}\left(\mathbf{x}_0^{(n)}, \mathbf{x}_t, t\right) = \mathbf{x}_0^{(n)} ,\quad \text{for large enough }  n .\]
Let \(\mathbf{x}_0^{*} = \mathbf{x}_0^{(n)} \) as \(n\) large enough. 
Next, from the expression that
\[\mathbf{f}_{\boldsymbol{\theta}}\left(\mathbf{x}_0^{(n)}, \mathbf{x}_t, t\right) =\mathbf{x}_t-\mathbf{F}_{\boldsymbol{\theta}}\left(\mathbf{x}_0^{(n)}, \mathbf{x}_t, t\right)\] 
we have:
\[\mathbf{x}_0^{*}=\mathbf{x}_t-\mathbf{F}_{\boldsymbol{\theta}}\left(\mathbf{x}_0^{(n)}, \mathbf{x}_t, t\right)\]
Further, we integral both sides of \(\frac{\mathrm{d} \mathbf{x}_s}{\mathrm{~d} s}=-\mathbf{h}_{\boldsymbol{\theta}}\left(\mathbf{x}_0^{(n)}, \mathbf{x}_s, s\right)\) and obtain:

\[
\begin{aligned}
\mathbf{x}_0 &=\mathbf{x}_t-\int_t^0 \mathbf{h}_{\boldsymbol{\theta}}\left(\mathbf{x}_0^{(n)}, \mathbf{x}_s, s\right) \mathrm{d} s \\&
=\mathbf{x}_t-\mathbf{F}_{\boldsymbol{\theta}}\left(\mathbf{x}_0^{(n)}, \mathbf{x}_t, t\right).
\end{aligned}
\]
By Theorem 1, the solution of the IVP is unique, which leads to \(\mathbf{x}_0^{*} = \mathbf{x}_0\).

Therefore, we obtain: 
\[\mathbf{f}_{\boldsymbol{\theta}}\left(\mathbf{x}_0^{(n)}, \mathbf{x}_t, t\right)\rightarrow \mathbf{x}_0,  \text{ as } n \text{ sufficiently large.}\]
and complete the proof.

Proof of \textbf{Theorem 3}:

Consider the IVP \Cref{ivp:2}: 
\[
\left\{
\begin{aligned}
    \frac{\mathrm{~d} \Tilde{\mathbf{x}}_s}{\mathrm{~d} s}&=\mathbf{h}_{\boldsymbol{\theta}^*}(\Tilde{\mathbf{x}}_t^{(n)}, \Tilde{\mathbf{x}}_s, s) && \quad s \in [0, t] \\
    \Tilde{\mathbf{x}}_0 &= \hat{\mathbf{x}}_t
\end{aligned}
\right.
\]

From the Lipschitz condition on $\mathbf{h}_{\boldsymbol{\theta}}$, we have:
\[
\left\|\mathbf{h}_{\boldsymbol{\theta}^*}(\tilde{\mathbf{x}}_t^{(n)}, \tilde{\mathbf{x}}_s, s)-\mathbf{h}_{\boldsymbol{\theta}^*}(\tilde{\mathbf{y}}_t^{(n)}, \tilde{\mathbf{y}}_s, s)\right\|_2 \leq L\left\|\tilde{\mathbf{x}}_s-\tilde{\mathbf{y}}_s\right\|_2 .
\]

Use the integral form:
\[
\left\|\mathbf{f}_{\boldsymbol{\theta}^*}\left(\mathbf{x}_0^{(n)}, \mathbf{x}_t, t\right)-\mathbf{f}_{\boldsymbol{\theta}^*}\left(\mathbf{y}_0^{(n)}, \mathbf{y}_t, t\right)\right\|_2 \leq \left\|\tilde{\mathbf{x}}_0-\tilde{\mathbf{y}}_0\right\|_2 + \int_0^t L\left\|\tilde{\mathbf{x}}_s-\tilde{\mathbf{y}}_s\right\|_2  d s
\]

By using Gröwnwall inequality, we have:
\[
\left\|\mathbf{f}_{\boldsymbol{\theta}^*}\left(\mathbf{x}_0^{(n)}, \mathbf{x}_t, t\right)-\mathbf{f}_{\boldsymbol{\theta}^*}\left(\mathbf{y}_0^{(n)}, \mathbf{y}_t, t\right)\right\|_2 \leq e^{Lt}\left\|\tilde{\mathbf{x}}_0-\tilde{\mathbf{y}}_0\right\|_2 = e^{Lt}\left\|\mathbf{x}_t-\mathbf{y}_t\right\|_2
\]
Next, consider the inverse time ODE \Cref{ivp:1}, we have: 
\[
\left\|\mathbf{x}_t-\mathbf{y}_t\right\|_2 \leq \left\|\mathbf{f}_{\boldsymbol{\theta}^*}(\mathbf{x}_0^{(n)}, \mathbf{x}_t, t)-\mathbf{f}_{\boldsymbol{\theta}^*}(\mathbf{y}_0^{(n)}, \mathbf{y}_t, t)\right\|_2 +\int_0^t L\left\|\mathbf{x}_s-\mathbf{y}_s\right\|_2 d s\]
Again, by using Gröwnwall inequality,
\[\left\|\mathbf{x}_t-\mathbf{y}_t\right\|_2 \leq e^{Lt}\left\|\mathbf{f}_{\boldsymbol{\theta}^*}(\mathbf{x}_0^{(n)}, \mathbf{x}_t, t)-\mathbf{f}_{\boldsymbol{\theta}^*}(\mathbf{y}_0^{(n)}, \mathbf{y}_t, t)\right\|_2 \]
Therefore,
\[\left\|\mathbf{f}_{\boldsymbol{\theta}^*}\left(\mathbf{x}_0^{(n)}, \mathbf{x}_t, t\right)-\mathbf{f}_{\boldsymbol{\theta}^*}\left(\mathbf{y}_0^{(n)}, \mathbf{y}_t, t\right)\right\|_2 \geq e^{-L t}\left\|\mathbf{x}_t-\mathbf{y}_t\right\|_2 \]

and we complete the proof of:
\[
e^{-Lt}\left\|\mathbf{x}_t-\mathbf{y}_t\right\|_2 \leq \left\|\mathbf{f}_{\boldsymbol{\theta}^*}(\mathbf{x}_0^{(n)},\mathbf{x}_t, t)-\mathbf{f}_{\boldsymbol{\theta}^*}(\mathbf{y}_0^{(n)},\mathbf{y}_t, t)\right\|_2 \leq e^{Lt}\left\|\mathbf{x}_t-\mathbf{y}_t\right\|_2 .
\]

\newpage

\begin{figure}[t]
    \begin{subfigure}{\textwidth}
        \centering
        \includegraphics[]{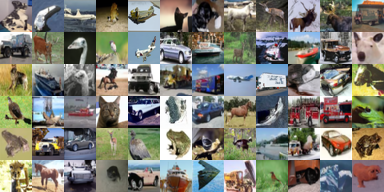}
        \caption{One-step samples from the uDDDM-VE model (FID=2.91).}
    \end{subfigure}
    \begin{subfigure}{\textwidth}
        \centering
        \includegraphics[]{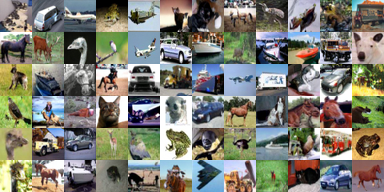}
        \caption{Two-step samples from the uDDDM-VE model (FID=2.68).}
    \end{subfigure}
    \begin{subfigure}{\textwidth}
        \centering
        \includegraphics[]{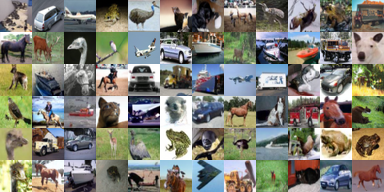}
        \caption{1000-step samples from the uDDDM-VE model (FID=1.89).}
    \end{subfigure}
    \caption{Uncurated samples from the uDDDM-VE model. All corresponding samples use the same initial noise.}
\end{figure}

\begin{figure}[t]
    \begin{subfigure}{\textwidth}
        \centering
        \includegraphics[]{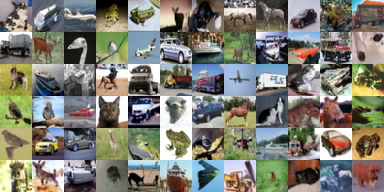}
        \caption{One-step samples from the uDDDM-VE-deep model (FID=2.63).}
    \end{subfigure}
    \begin{subfigure}{\textwidth}
        \centering
        \includegraphics[]{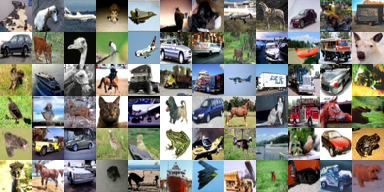}
        \caption{Two-step samples from the uDDDM-VE-deep model (FID=2.35).}
    \end{subfigure}
    \begin{subfigure}{\textwidth}
        \centering
        \includegraphics[]{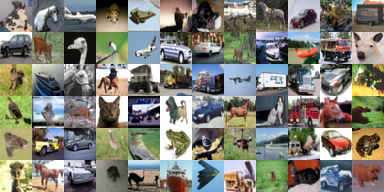}
        \caption{1000-step samples from the uDDDM-VE-deep model (FID=1.65).}
    \end{subfigure}
    \caption{Uncurated samples from the uDDDM-VE-deep model. All corresponding samples use the same initial noise.}
\end{figure}

\begin{figure}[t]
    \begin{subfigure}{\textwidth}
        \centering
        \includegraphics[]{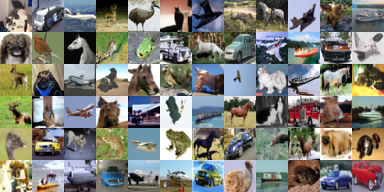}
        \caption{One-step samples from the uDDDM-VP model (FID=2.84).}
    \end{subfigure}
    \begin{subfigure}{\textwidth}
        \centering
        \includegraphics[]{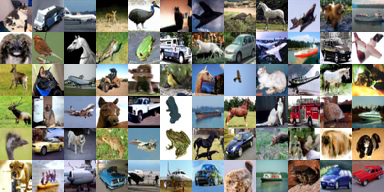}
        \caption{Two-step samples from the uDDDM-VP model (FID=2.50).}
    \end{subfigure}
    \begin{subfigure}{\textwidth}
        \centering
        \includegraphics[]{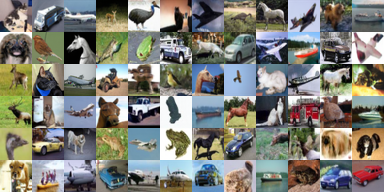}
        \caption{1000-step samples from the uDDDM-VP model (FID=1.73).}
    \end{subfigure}
    \caption{Uncurated samples from the uDDDM-VP model. All corresponding samples use the same initial noise.}
\end{figure}

\begin{figure}[t]
    \begin{subfigure}{\textwidth}
        \centering
        \includegraphics[]{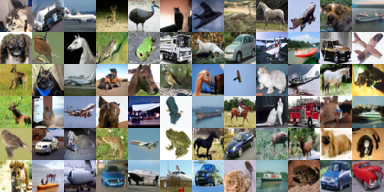}
        \caption{One-step samples from the uDDDM-VP-deep model (FID=2.53).}
    \end{subfigure}
    \begin{subfigure}{\textwidth}
        \centering
        \includegraphics[]{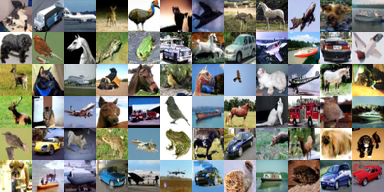}
        \caption{Two-step samples from the uDDDM-VP-deep model (FID=2.21).}
    \end{subfigure}
    \begin{subfigure}{\textwidth}
        \centering
        \includegraphics[]{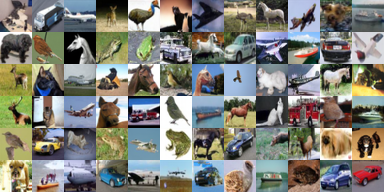}
        \caption{1000-step samples from the uDDDM-VP-deep model (FID=1.65).}
    \end{subfigure}
    \caption{Uncurated samples from the uDDDM-VP-deep model. All corresponding samples use the same initial noise.}
\end{figure}

\end{document}